\documentclass[twoside,twocolumn]{article}

\usepackage{graphicx}
\usepackage{algorithm}
\usepackage{algorithmic}
\usepackage{amsmath}
\usepackage{txfonts}
\usepackage[all]{xy}
\newtheorem{defn}{Definition}
\newtheorem{lem}{Lemma}
\newtheorem{thm}{Theorem}
\newtheorem{proposition}{Proposition}
\newenvironment{keywords}{\vskip 1ex\noindent{\bf Keywords:}}{}
\def\paradot#1{\paragraph{#1.}}
\newcommand{\cdbar}{\,|\,}
\newcommand{\cbar}{\,|\,}
\newcommand{\cX}{{\cal X}}
\newcommand{\cA}{{\cal A}}
\newcommand{\cO}{{\cal O}}
\newcommand{\cR}{{\cal R}}
\newcommand{\searchalg}{{$\rho$UCT}}
\newcommand{\bstr}[1]{\llbracket #1 \rrbracket}

\begin{document}

\title{\vspace{-4ex}
\vskip 2mm\bf\LARGE\hrule height5pt \vskip 4mm
Reinforcement Learning via AIXI Approximation
\vskip 4mm \hrule height2pt}

\author{\normalsize\hspace{-2ex}
\begin{minipage}{0.26\textwidth}
\begin{center}
Joel Veness \\
University of NSW \& NICTA \\
\small{{\sc joelv@cse.unsw.edu.au}}
\end{center}
\end{minipage}
\hfil
\begin{minipage}{0.24\textwidth}
\begin{center}
Kee Siong Ng \\
Medicare Australia \& ANU \\
\small{{\sc keesiong.ng@gmail.com}}
\end{center}
\end{minipage}
\hfil
\begin{minipage}{0.24\textwidth}
\begin{center}
Marcus Hutter \\
ANU \& NICTA \\
\small{{\sc marcus.hutter@anu.edu.au}}
\end{center}
\end{minipage}
\hfil
\begin{minipage}{0.24\textwidth}
\begin{center}
David Silver \\
University College London \\
\small{{\sc davidstarsilver@googlemail.com}}
\end{center}
\end{minipage}
}

\date{13 July 2010}

\maketitle

\def\abstractname{\centering\large Abstract}
\begin{abstract}
This paper introduces a principled approach for the design of a scalable general reinforcement learning agent.
This approach is based on a direct approximation of AIXI, a Bayesian optimality notion for general reinforcement learning agents.
Previously, it has been unclear whether the theory of AIXI could motivate the design of practical algorithms.
We answer this hitherto open question in the affirmative, by providing the first computationally feasible approximation to the AIXI agent.
To develop our approximation, we introduce a Monte Carlo Tree Search algorithm along with an agent-specific extension of the Context Tree Weighting algorithm.
Empirically, we present a set of encouraging results on a number of stochastic, unknown, and partially observable domains.
\def\contentsname{\centering\normalsize Contents}
{\parskip=-2.5ex\tableofcontents}
\end{abstract}


\section{Introduction}

Consider an agent that exists within some unknown environment.
The agent interacts with the environment in cycles.
At each cycle, the agent executes an action and receives in turn an observation and a reward.
The \emph{general reinforcement learning problem} is to construct an agent that, over time, collects as much reward as
possible from an initially {\em unknown} environment.

The AIXI agent \cite{Hutter:04uaibook} is a formal, mathematical solution to the general reinforcement learning problem.
It can be decomposed into two main components: planning and prediction.
Planning amounts to performing an expectimax operation to determine each action.
Prediction uses Bayesian model averaging, over the largest possible model class expressible on a Turing Machine, to predict future observations and rewards based on past experience.
AIXI is shown in \cite{Hutter:04uaibook} to be optimal in the sense that it will rapidly learn an accurate model of the unknown environment and exploit it to maximise its expected future reward.

As AIXI is only asymptotically computable, it is by no means an algorithmic solution to the general reinforcement learning problem.
Rather it is best understood as a Bayesian \emph{optimality notion} for decision making in general unknown environments.
This paper demonstrates, for the first time, how a practical agent can be built from the AIXI theory.
Our solution directly approximates the planning and prediction components of
AIXI.
In particular, we use a generalisation of UCT \cite{kocsis06} to approximate
the expectimax operation, and an agent-specific extension of CTW \cite{ctw95},
a Bayesian model averaging algorithm for prediction suffix trees, for prediction and learning.
Perhaps surprisingly, this kind of direct approximation is possible, practical and theoretically appealing.
Importantly, the essential characteristic of AIXI, its generality, can be largely preserved.

\section{The Agent Setting}\label{sec:agent setting}

This section introduces the notation and terminology we will use to describe strings of agent experience, the true underlying environment and the agent's model of the environment.

The (finite) action, observation, and reward spaces are denoted by $\cA, \cO$, and $\cR$ respectively.
An observation-reward pair $or$ is called a percept.
We use ${\cal X}$ to denote the percept space $\cO \times \cR$.

\begin{defn}
A history $h$ is an element of $({\cal A} \times {\cal X})^* \cup ({\cal A} \times {\cal X})^* \times {\cal A}$.
\end{defn}

\noindent {\textbf Notation:} A string $x_1x_2 \ldots x_n$ of length $n$ is denoted by $x_{1:n}$.
The empty string is denoted by $\epsilon$.
The concatenation of two strings $s$ and $r$ is denoted by $sr$.
The prefix $x_{1:j}$ of $x_{1:n}$, $j\leq n$, is denoted by $x_{\leq j}$ or $x_{< j+1}$.
The notation generalises for blocks of symbols: e.g.\ $ax_{1:n}$ denotes $a_1x_1a_2x_2\ldots a_nx_n$ and $ax_{<j}$ denotes $a_1x_1 a_2x_2\ldots a_{j-1}x_{j-1}$.

The following definition states that the environment takes the form of a probability distribution over possible percept sequences
conditioned on actions taken by the agent.

\begin{defn}
\label{def:environment_model}
An environment $\rho$ is a sequence of conditional probability functions $\{ \rho_0, \rho_1, \rho_2, \dots \}$,
where $\rho_n \colon \mathcal{A}^n \rightarrow \text{Density}\;(\mathcal{X}^n)$, that satisfies
\begin{equation}\label{eq:chronological}
\forall a_{1:n} \forall x_{<n}: \,\rho_{n-1}(x_{<n} \cbar a_{<n}) = \sum_{x_n \in {\cal X}}\, \rho_n(x_{1:n} \cbar a_{1:n}).
\end{equation}
In the base case, we have $\rho_0(\epsilon \cbar \epsilon) = 1$.
\end{defn}

Equation \ref{eq:chronological}, called the chronological condition in \cite{Hutter:04uaibook}, captures the natural constraint that action $a_n$
has no effect on observations made before it.
For convenience, we drop the index $n$ in $\rho_n$ from here onwards.

Given an environment $\rho$,
\begin{equation}
 \rho( x_n \cbar ax_{<n}a_n ) := \frac{ \rho (x_{1:n} \cbar a_{1:n}) }{ \rho( x_{<n} \cbar a_{<n} ) }\label{cond prob of or}
\end{equation}
is the $\rho$-probability of observing $x_n$ in cycle $n$ given history $h=ax_{<n}a_n$,
provided $\rho( x_{<n} \cbar a_{<n} ) > 0$.
It now follows that
\begin{equation}
\label{eq:ChainRule}
\rho(x_{1:n} \cbar a_{1:n}) = \rho( x_1 \cbar a_1) \rho( x_2 \cbar ax_1a_2) \cdots \rho( x_n \cbar ax_{<n}a_n).
\end{equation}

Definition \ref{def:environment_model} is used to describe both the true (but unknown) underlying environment
and the agent's \emph{subjective} model of the environment.
The latter is called the agent's \emph{environment model} and is typically learnt from data.
Definition \ref{def:environment_model} is extremely general.
It captures a wide variety of environments, including standard reinforcement learning setups such as MDPs and POMDPs.

The agent's goal is to accumulate as much reward as it can during its lifetime.
More precisely, the agent seeks a \emph{policy} that will allow it to maximise its expected future reward up to a fixed, finite, but
arbitrarily large horizon $m \in \mathbb{N}$.
Formally, a policy is a function that maps a history to an action.
The expected future value of an agent acting under a particular policy is defined as follows.

\begin{defn}
Given history $ax_{1:t}$, the $m$-horizon expected future reward of an agent acting under policy $\pi \colon (\mathcal{A} \times \mathcal{X})^* \rightarrow \cA$ with respect to an environment $\rho$ is
\begin{equation}\label{eq:val}
v_{\rho}^m(\pi, ax_{1:t}) := \mathbb{E}_{x_{t+1:t+m} \sim \rho} \left[ \sum\limits_{i=t+1}^{t+m} R_i(ax_{\leq t+m} )  \right],
\end{equation}
where for $t + 1 \leq k \leq t+m$, $a_k := \pi(ax_{<k})$, and $R_k(aor_{\leq t+m}) := r_k$.
The quantity $v_{\rho}^m(\pi, ax_{1:t}a_{t+1})$ is defined similarly, except that $a_{t+1}$ is now no longer defined by $\pi$.
\end{defn}

The optimal policy $\pi^*$ is the policy that maximises the expected future reward.
The maximal achievable expected future reward of an agent with history $h$ in environment $\rho$ looking $m$ steps ahead is $V_{\rho}^m(h) := v_{\rho}^m(\pi^*, h)$.
It is easy to see that if $h\equiv ax_{1:t} \in ({\cal A} \times {\cal X})^{t}$, then
{\scriptsize\begin{equation}\label{value defn}
V_{\rho}^{m}(h) = \max\limits_{a_{t+1}}\sum\limits_{x_{t+1}}{\rho(x_{t+1} \cbar h a_{t+1})} \cdots
\max\limits_{a_{t+m}}\sum\limits_{x_{t+m}}{\rho(x_{t+m} \cbar h ax_{t+1:t+m-1} a_{t+m})}
\left[\sum\limits_{i=t+1}^{t+m}r_i\right].
\end{equation}}

We will refer to Equation \ref{value defn} as the \emph{expectimax operation}.
The $m$-horizon optimal action $a^*_{t+1}$ at time $t+1$ is related to the expectimax operation by
\begin{equation}
a^*_{t+1} = \arg\max\limits_{a_{t+1}}V_{\rho}^{m}(ax_{1:t}a_{t+1})
\end{equation}

Eqs \ref{eq:val} and \ref{value defn} can be modified to handle discounted reward, however we focus on the finite-horizon case since it both aligns with AIXI and allows for a simplified presentation.

\section{Bayesian Agents}\label{sec:BayesAgents}

In the general reinforcement learning setting, the environment $\rho$ is unknown to the agent.
One way to learn an environment model is to take a Bayesian approach.
Instead of committing to any single environment model, the agent uses a \emph{mixture} of environment models.
This requires committing to a class of possible environments (the model class), assigning an initial weight to each possible environment (the prior), and subsequently updating the weight for each model using Bayes rule (computing the posterior) whenever more experience is obtained.

The above procedure is similar to Bayesian methods for predicting sequences
of (singly typed) observations.
The key difference in the agent setup is that each prediction is now also
dependent on previous agent actions.
We incorporate this by using the \emph{action-conditional} definitions and identities of Section \ref{sec:agent setting}.

\begin{defn}
\label{def:mixture_environment}
Given a model class $\mathcal{M} := \{ \rho_1, \rho_2, \dots\}$ and a prior weight $w_0^{\rho}>0$ for each $\rho \in \mathcal{M}$ such that $\sum_{\rho\in\mathcal{M}}w_0^\rho=1$,
the mixture environment model is  $\xi(x_{1:n} \cbar a_{1:n}) := \sum\limits_{\rho \in \mathcal{M}} w_0^{\rho} \rho(x_{1:n} \cbar a_{1:n})$.
\end{defn}

The next result follows immediately.

\begin{proposition}
\label{prop:MixtureEnvModelsAreEnvModels}
A mixture environment model is an environment model.
\end{proposition}

Proposition~\ref{prop:MixtureEnvModelsAreEnvModels} allows us to use a mixture environment model whenever we can use an environment model.
Its importance will become clear shortly.

To make predictions using a mixture environment model $\xi$, we use
\begin{equation}
\label{eq:MixturePredictor}
\xi(x_n \cbar ax_{<n}a_n) = \frac{\xi(x_{1:n} \cbar a_{1:n})}{\xi(x_{<n} \cbar a_{<n})},
\end{equation}
which follows from Proposition~\ref{prop:MixtureEnvModelsAreEnvModels} and Eq.~\ref{cond prob of or}.
The RHS of Eq.~\ref{eq:MixturePredictor} can be written out as a convex combination of model predictions to give
\begin{equation}
\label{eq:PosteriorPredictor}
\xi(x_n \cbar ax_{<n}a_n) = \sum\limits_{\rho \in \mathcal{M}} w_{n-1}^{\rho} \rho(x_n \cbar ax_{<n}a_n),
\end{equation}
where the posterior weight $w_{n-1}^{\rho}$ for $\rho$ is given by
\begin{equation}
\label{eq:PosteriorWeight}
w_{n-1}^{\rho} := \frac{w_0^{\rho} \rho(x_{<n} \cbar a_{<n})}{\sum\limits_{\mu \in \mathcal{M}} w_0^{\mu} \mu(x_{<n} \cbar a_{<n})} = \Pr(\rho \cbar ax_{<n}).
\end{equation}

Bayesian agents enjoy a number of strong theoretical performance guarantees; these are explored in Section \ref{sec:theory}.
In practice, the main difficulty in using a mixture environment model is computational.
A rich model class is required if the mixture environment model is to possess general prediction capabilities, however naively using (\ref{eq:PosteriorPredictor}) for online prediction requires at least $O(|\mathcal{M}|)$ time to process each new piece of experience.
One of our main contributions, introduced in Section \ref{sec:ctw}, is a large, efficiently computable mixture environment model that runs in time $O(\log(\log |\mathcal{M}|))$.
Before looking at that, we will examine in the next section a Monte Carlo Tree Search algorithm for approximating the expectimax operation.

\section{Monte Carlo Expectimax Approximation}\label{sec:mcts}

Full-width computation of the expectimax operation (\ref{value defn}) takes $O(|\cA \times \cX|^m)$ time, which is unacceptable for all but tiny values of $m$.
This section introduces \searchalg, a generalisation of the popular UCT algorithm \cite{kocsis06} that can be used to approximate a finite horizon expectimax operation given an environment model $\rho$.
The key idea of Monte Carlo search is to sample observations from the environment, rather than exhaustively considering all possible observations.
This allows for effective planning in environments with large observation spaces.
Note that since an environment model subsumes both MDPs and POMDPs, \searchalg\ effectively extends the UCT algorithm to a wider class of problem domains.

The UCT algorithm has proven effective in solving large discounted or finite horizon MDPs.
It assumes a generative model of the MDP that when given a state-action pair $(s, a)$ produces a subsequent state-reward pair $(s', r)$ distributed according to $\Pr(s',r \cbar s, a)$.
By successively sampling trajectories through the state space, the UCT algorithm incrementally constructs a search tree, with each node containing an estimate of the value of each state.
Given enough time, these estimates converge to the true values.

The \searchalg\ algorithm can be realised by replacing the notion of state in UCT by an agent history $h$ (which is always a sufficient statistic) and using an environment model $\rho(or \cbar h)$ to predict the next percept.
The main subtlety with this extension is that the history used to determine the conditional probabilities must be updated \emph{during the search} to reflect the extra information an agent will have at a hypothetical future point in time.

We will use $\Psi$ to represent all the nodes in the search tree, $\Psi(h)$ to represent the node corresponding to a particular history $h$, $\hat{V}_\rho^m(h)$ to represent the sample-based estimate of the expected future reward, and $T(h)$ to denote the number of times a node $\Psi(h)$ has been sampled.
Nodes corresponding to histories that end or do not end with an action are called chance and decision nodes respectively.

Algorithm \ref{alg:rhoUCT} describes the top-level algorithm, which the agent
calls at the beginning of each cycle.
It is initialised with the agent's total experience $h$ (up to time $t$) and the planning horizon $m$.
It repeatedly invokes the {\sc Sample} routine until out of time.
Importantly, \searchalg\ is an \emph{anytime} algorithm; an approximate best action, whose quality improves with time, is always available.
This is retrieved by {\sc BestAction}, which computes
$a^*_{t} = \arg\max\limits_{a_{t}}\hat{V}_{\rho}^{m}(ax_{<t}a_{t})$.

\algsetup{indent=2em}
\begin{algorithm}[h!]
\caption{\searchalg$(h, m)$}\label{alg:rhoUCT}
\begin{algorithmic}[1]
\REQUIRE A history $h$
\REQUIRE A search horizon $m \in \mathbb{N}$
\medskip
\STATE {\sc Initialise}$(\Psi)$
\REPEAT
  \STATE {\sc Sample}$(\Psi, h, m)$
\UNTIL {out of time}
\RETURN {\sc BestAction}$(\Psi,h)$
\end{algorithmic}
\end{algorithm}

Algorithm \ref{alg:sample} describes the recursive routine used to sample a single future trajectory.
It uses the {\sc SelectAction} routine to choose moves at interior nodes, and invokes the {\sc Rollout} routine at unexplored leaf nodes.
The {\sc Rollout} routine picks actions uniformly at random until the (remaining) horizon is reached, returning the accumulated reward.
After a complete trajectory of length $m$ is simulated, the value estimates are updated for each node traversed.
Notice that the recursive calls on Lines \ref{alg:sample:rec1} and \ref{alg:sample:rec2} append the most recent percept or action to the history argument.

\algsetup{indent=2em}
\begin{algorithm}
\caption{{\sc Sample}$(\Psi, h, m)$}\label{alg:sample}
\begin{algorithmic}[1]
\REQUIRE A search tree $\Psi$
\REQUIRE A history $h$
\REQUIRE A remaining search horizon $m \in \mathbb{N}$
\medskip
\IF {$m=0$}
  \RETURN 0
\ELSIF {$\Psi(h)$ is a chance node}
  \STATE  Generate $(o,r)$ from $\rho(or \cbar h)$
  \STATE  Create node $\Psi(hor)$ if $T(hor) = 0$
  \STATE  reward $\leftarrow$ $r$ $+$ {\sc Sample}$(\Psi, hor, m - 1)$\label{alg:sample:rec1}
\ELSIF {$T(h)=0$}
  \STATE reward $\leftarrow$ {\sc Rollout}$(h, m)$
\ELSE
  \STATE $a$ $\leftarrow$ {\sc SelectAction}$(\Psi, h, m)$
  \STATE reward $\leftarrow$ {\sc Sample}$(\Psi, ha, m)$\label{alg:sample:rec2}
\ENDIF
\STATE $\hat{V}(h) \leftarrow \frac{1}{T(h)+1}[reward + T(h)\hat{V}(h)]$
\STATE $T(h) \leftarrow T(h)+1$
\RETURN reward
\end{algorithmic}
\end{algorithm}

Algorithm \ref{alg:select_action} describes the UCB \cite{auer02} policy used to select actions at decision nodes.
The $\alpha$ and $\beta$ constants denote the smallest and largest elements of $\cR$ respectively.
The parameter $C$ varies the selectivity of the search; larger values grow bushier trees.
UCB automatically focuses attention on the best looking action in such a way that the sample estimate $\hat{V}_\rho(h)$ converges to $V_\rho(h)$, whilst still exploring alternate actions sufficiently often to guarantee that the best action will be found.

The ramifications of the \searchalg\ extension are particularly significant to Bayesian agents described in Section \ref{sec:BayesAgents}.
~Proposition~\ref{prop:MixtureEnvModelsAreEnvModels} allows \searchalg\ to be instantiated with a mixture environment model, which directly incorporates model uncertainty into the planning process.
This gives (in principle, provided that the model class contains the true environment and ignoring issues of limited computation) the well known Bayes-optimal solution to the exploration/exploitation dilemma;
namely, if a reduction in model uncertainty would lead to higher expected future reward, \searchalg\ would recommend an information gathering action.

\algsetup{indent=2em}
\begin{algorithm}[h!]
\caption{{\sc SelectAction}$(\Psi, h, m)$}\label{alg:select_action}
\begin{algorithmic}[1]
\REQUIRE A search tree $\Psi$
\REQUIRE A history $h$
\REQUIRE A remaining search horizon $m \in \mathbb{N}$
\REQUIRE An exploration/exploitation constant $C > 0$
\medskip
\STATE $\mathcal{U} = \{ a\in \cA \colon T(ha) = 0\}$
\IF {$\mathcal{U} \ne \{\}$}
  \STATE Pick $a \in \mathcal{U}$ uniformly at random
  \STATE Create node $\Psi(ha)$
  \RETURN a
\ELSE
  \RETURN $\arg\max\limits_{a \in \cA} \left\{ \frac{1}{m(\beta - \alpha)} \hat{V}(ha) + C \sqrt{\frac{\log(T(h))}{T(ha)}} \right\}$
\ENDIF
\end{algorithmic}
\end{algorithm}

\section{Action-Conditional CTW}\label{sec:ctw}

We now introduce a large mixture environment model for use with \searchalg.
Context Tree Weighting (CTW) \cite{ctw95} is an efficient and theoretically well-studied binary sequence prediction
algorithm that works well in practice.
It is an online Bayesian model averaging algorithm that computes, at each time point $t$, the probability
\begin{equation}\label{eq:ctw pr} \Pr (y_{1:t}) = \sum_{M} \Pr(M) \Pr( y_{1:t} \cbar M), \end{equation}
where $y_{1:t}$ is the binary sequence seen so far, $M$ is a prediction suffix tree \cite{ron96},
$\Pr(M)$ is the prior probability of $M$, and the summation is over \emph{all} prediction suffix trees of bounded depth $D$.
A naive computation of (\ref{eq:ctw pr}) takes time $O(2^{2^D})$; using CTW, this computation requires only $O(D)$ time.
In this section, we outline how CTW can be extended to compute probabilities of the form
\begin{equation}\label{eq:gen ctw pr} \Pr ( x_{1:t} \cbar a_{1:t}) = \sum_{M} \Pr(M) \Pr( x_{1:t} \cbar M, a_{1:t}), \end{equation}
where $x_{1:t}$ is a percept sequence, $a_{1:t}$ is an action sequence, and $M$ is a prediction suffix tree
as in (\ref{eq:ctw pr}).
This extension allows CTW to be used as a mixture environment model (Definition~\ref{def:mixture_environment}) in
the \searchalg\ algorithm, where we combine (\ref{eq:gen ctw pr}) and (\ref{cond prob of or}) to predict the next percept given a history.

\paradot{Krichevsky-Trofimov Estimator}
We start with a brief review of the KT estimator for Bernoulli distributions.
Given a binary string $y_{1:t}$ with $a$ zeroes and $b$ ones, the KT estimate of the probability of the next symbol is given by
\begin{gather}
\text{Pr}_{kt}( Y_{t+1} = 1 \cbar y_{1:t} ) := \frac{b + 1/2}{a + b + 1}. \label{kt update}
\end{gather}
The KT estimator can be obtained via a Bayesian analysis by putting an uninformative (Jeffreys Beta(1/2,1/2)) prior $\text{Pr}(\theta)\propto\theta^{-1/2}(1-\theta)^{-1/2}$ on the parameter $\theta\in[0,1]$ of the Bernoulli distribution.
The probability of a string $y_{1:t}$ is given by
\begin{align*}
  \text{Pr}_{kt}( y_{1:t} ) &= \text{Pr}_{kt}(y_1 \cbar \epsilon) \text{Pr}_{kt}(y_2 \cbar y_1) \cdots \text{Pr}_{kt} (y_t \cbar y_{<t}) \\
  &= \textstyle{\int} \theta^b(1-\theta)^a\,\text{Pr}(\theta)\,d\theta.
\end{align*}

\paradot{Prediction Suffix Trees}
We next describe prediction suffix trees.
We consider a binary tree where all the left edges are labelled 1 and all the right edges
are labelled 0.
The depth of a binary tree $M$ is denoted by $d(M)$.
Each node in $M$ can be identified by a string in $\{0,1\}^*$ as usual: $\epsilon$ represents the root node of $M$;
and if $n \in \{0,1\}^*$ is a node in $M$, then $n1$ and $n0$ represent respectively the
left and right children of node $n$.
The set of $M$'s leaf nodes is denoted by $L(M) \subset \{0,1\}^*$.
Given a binary string $y_{1:t}$ where $t \geq d(M)$,
we define $M(y_{1:t}) := y_ty_{t-1}\ldots y_{t'}$, where $t' \leq t$ is the (unique) positive integer such that
$y_{t}y_{t-1}\ldots y_{t'} \in L(M)$.

\begin{defn}\label{defn:pst}
A prediction suffix tree (PST) is a pair $(M,\Theta)$, where $M$ is a binary tree and
associated with each $l \in L(M)$ is a distribution over $\{0,1\}$ parameterised by $\theta_l \in \Theta$.
We call $M$ the model of the PST and $\Theta$ the parameter of the PST.
\end{defn}

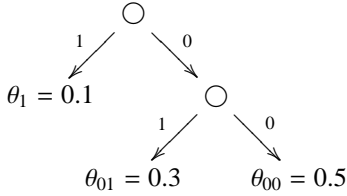
\begin{figure}[!htbp]
\hspace{2em}\centerline{
\xymatrix @ur {
\theta_1 = 0.1  & {\bigcirc} \ar[l]_1 \ar[d]^0 \\
\theta_{01} = 0.3 & \bigcirc \ar[l]_1 \ar[d]^0 \\
  & \theta_{00} = 0.5 & \hspace*{5em} }}
\vspace{-3.2em}
\caption{An example prediction suffix tree}\label{fig:pst}
\end{figure}

A PST $(M,\Theta)$ maps each binary string $y_{1:t}$, $t \geq d(M)$, to $\theta_{M(y_{1:t})}$;
the intended meaning is that $\theta_{M(y_{1:t})}$ is the probability that the next bit following $y_{1:t}$ is 1.
For example, the PST in Figure~\ref{fig:pst} maps the string 1110 to $\theta_{M(1110)} = \theta_{01} = 0.3$, which means the
next bit after 1110 is 1 with probability 0.3.

\paradot{Action-Conditional PST}
In the agent setting, we reduce the problem of predicting history sequences with general non-binary alphabets to that of predicting the
bit representations of those sequences.
Further, we only ever condition on actions; this is achieved by appending bit representations of actions to the input sequence without updating the PST parameters.

Assume $|{\cal X}| = 2^{l_{\cal X}}$ for some $l_{\cal X} > 0$.
Denote by $\bstr{x} = x[1,l_{\cal X}] = x[1]x[2]\ldots x[l_{\cal X}]$
the bit representation of $x \in {\cal X}$.
Denote by $\bstr{x_{1:t}} = \bstr{x_1}\bstr{x_2}\ldots\bstr{x_t}$ the bit representation of a sequence $x_{1:t}$.
Action symbols are treated similarly.

To do action-conditional sequence prediction using a PST with a given model $M$ but unknown parameter, we start with
$\theta_l := \Pr_{kt}(1 \cbar \epsilon)=1/2$ at each $l \in L(M)$.
We set aside an initial portion of the binary history sequence to initialise the variable $h$ and then
repeat the following steps as long as needed:
\begin{enumerate}\itemsep1mm\parskip0mm
 \item set $h := h\bstr{a}$, where $a$ is the current selected action;
 \item for $i := 1$ to $l_{\cal X}$ do
  \begin{enumerate}\itemsep1mm\parskip0mm
    \item predict the next bit using the distribution $\theta_{M(h)}$;
    \item observe the next bit $x[i]$, update $\theta_{M(h)}$ using (\ref{kt update}) according to the value of $x[i]$, and then set $h := hx[i]$.
  \end{enumerate}
\end{enumerate}

Let $M$ be the model of a prediction suffix tree, $a_{1:t}$ an action sequence,
$x_{1:t}$ a percept sequence, and $h := \bstr{ax_{1:t}}$.
For each node $n$ in $M$, define $h_{M,n}$ by
\begin{equation}\label{eq:bits at a node} h_{M,n} := h_{i_1} h_{i_2} \cdots h_{i_k}\end{equation}
where $1 \leq i_1 < i_2 < \cdots < i_k \leq t$ and, for each $i$, $i \in \{i_1,i_2,\ldots i_k\} \text{\;iff\;}
h_{i}$ is a percept bit and $n$ is a prefix of $M(h_{1:i-1})$.
We have the following expression for the probability of $x_{1:t}$ given $M$ and $a_{1:t}$:
\begin{align}
 \Pr( x_{1:t} \cdbar M, a_{1:t} )
  &=  \prod_{i=1}^t \prod_{j=1}^{l_{\cal X}} \Pr( x_i[j] \cbar M, \bstr{ax_{<i}a_i} x_i[1,j-1]) \notag \\
  &= \prod_{n \in L(M)} \text{Pr$_{kt}$} ( h_{M,n}). \label{tree prop1}
\end{align}

\paradot{Context Tree Weighting}
The above deals with action-conditional prediction using a single PST.
We now show how we can efficiently perform action-conditional prediction using a Bayesian mixture of PSTs.
There are two main computational tricks: the use of a data structure to represent all PSTs of a certain maximum depth
and the use of probabilities of sequences in place of conditional probabilities.

\begin{defn}\label{defn:context tree}
A context tree of depth $D$ is a perfect binary tree of depth $D$ such that
attached to each node (both internal and leaf) is a probability on $\{0,1\}^*$.
\end{defn}

The weighted probability $P^n_w$ of each node $n$ in the context tree $T$ after seeing $h := \bstr{ax_{1:t}}$ is defined as follows:
\begin{align*}
P^n_w := \begin{cases}
           \Pr_{kt} ( h_{T,n}) & \text{if $n$ is a leaf node;} \\
           \frac{1}{2}\Pr_{kt}( h_{T,n}) +
            \frac{1}{2}P^{n0}_w \times P^{n1}_w & \text{otherwise.}
          \end{cases}
\end{align*}

The following is a straightforward extension of a result due to \cite{ctw95}.
\begin{lem}
Let $T$ be the depth-$D$ context tree after seeing $h := \bstr{ax_{1:t}}$.
For each node $n$ in $T$ at depth $d$, we have
\begin{equation}
P^n_w  = \sum_{M \in C_{D-d}} 2^{-\Gamma_{D-d}(M)} \prod_{l \in L(M)} \text{{\em Pr}}_{kt}( h_{T,nl}),
\end{equation}
where $C_d$ is the set of all models of PSTs with depth $\leq d$, and $\Gamma_{d}(M)$ is the code-length for $M$ given
by the number of nodes in $M$ minus the number of leaf nodes in $M$ of depth $d$.
\label{model averaging}
\end{lem}
A corollary of Lemma \ref{model averaging} is that at the root node $\epsilon$ of the context tree $T$ after seeing $h := \bstr{ax_{1:t}}$,
we have
{\allowdisplaybreaks
\begin{align}
P^\epsilon_w(x_{1:t}|a_{1:t})
  &= \sum_{ M \in C_D} 2^{-\Gamma_{D}(M)} \prod_{l \in L(M)} \text{Pr}_{kt}( h_{T,l}) \label{corr 1}\\
  &= \sum_{ M \in C_D} 2^{-\Gamma_{D}(M)} \prod_{l \in L(M)} \text{Pr}_{kt}( h_{M,l}) \label{corr 2} \\
  &= \sum_{ M \in C_D} 2^{-\Gamma_{D}(M)} \Pr( x_{1:t} \cdbar M, a_{1:t}), \label{corr 3}
\end{align}
where} the last step follows from (\ref{tree prop1}).
Notice that the prior $2^{-\Gamma_D(\cdot)}$ penalises PSTs with large tree structures.
The conditional probability of $x_{t}$ given $ax_{<t}a_t$ can be obtained from (\ref{cond prob of or}).
We can also efficiently sample the individual bits of $x_t$ one by one.

\paradot{Computational Complexity}
The Action-Conditional CTW algorithm grows the context tree dynamically.
Using a context tree with depth $D$, there are at most $O(tD\log(|\cO||\cR|))$ nodes in the context tree after $t$ cycles.
In practice, this is a lot less than $2^D$, the number of nodes in a fully grown context tree.
The time complexity of Action-Conditional CTW is also impressive, requiring $O(D\log(|\cO||\cR|))$ time to process each new piece of agent experience and
$O(mD\log(|\cO||\cR|))$ to sample a single trajectory when combined with \searchalg.
Importantly, this is independent of $t$, which means that the computational overhead does not increase as the agent gathers more experience.

\section{Theoretical Results}\label{sec:theory}

Putting the \searchalg\ and Action-Conditional CTW algorithms together yields our approximate AIXI agent.
We now investigate some of its properties.

\paradot{Model Class Approximation}\label{subsec:model class apx}
By instantiating (\ref{value defn}) with the mixture environment model (\ref{corr 3}), one can show that the optimal action for an agent
at time $t$, having experienced $ax_{<t}$, is given by
{\small\begin{equation*}
\arg \max\limits_{a_t}\sum\limits_{x_t}\cdots \max\limits_{a_{t+m}}\sum\limits_{x_{t+m}} \left[ \sum\limits_{i=t}^{t+m}r_i \right]
\sum_{ M \in {\mathcal C}_{D}} 2^{-\Gamma_{D}(\rho)} \Pr( x_{1:t+m} \cdbar M, a_{1:t+m}). \label{aipsi}
\end{equation*}}
Compare this to the action chosen by the AIXI agent
\begin{equation*}
\label{aixi_eq2}
\arg\max\limits_{a_t}\sum\limits_{x_t} \dots \max\limits_{a_{t+m}} \sum\limits_{x_{t+m}}
 \left[ \sum\limits_{i=t}^{t+m}r_i \right]
\sum\limits_{\rho \in {\cal M}}2^{-K(\rho)} \rho( x_{1:t+m} \cbar a_{1:t+m}),
\end{equation*}
where class ${\cal M}$ consists of all computable environments $\rho$ and $K(\rho)$ denotes the Kolmogorov complexity of $\rho$.
Both use a prior that favours simplicity.
The main difference is in the subexpression describing the mixture over the model class.
AIXI uses a mixture over all enumerable chronological semimeasures, which is completely general but incomputable.
Our approximation uses a mixture of all prediction suffix trees of a certain maximum depth, which is still
a rather general class, but one that is efficiently computable.

\paradot{Consistency of \searchalg}
\cite{kocsis06} shows that the UCT algorithm is consistent in finite horizon MDPs and derive
finite sample bounds on the estimation error due to sampling.
By interpreting histories as Markov states, the general reinforcement learning problem reduces to a
finite horizon MDP and the results of \cite{kocsis06} are now directly applicable.
Restating the main consistency result in our notation, we have
\begin{equation}\label{uctc}
\forall \epsilon \forall h \lim_{T(h) \to \infty} \Pr \left( | V_{\rho}^m(h) - \hat{V}_\rho^m(h) | \leq \epsilon \right) = 1.
\end{equation}
Furthermore, the probability that a suboptimal action (with respect to $V^m_\rho(\cdot)$) is chosen by \searchalg\ goes to zero in the limit.

\paradot{Convergence to True Environment}\label{subsec:Upsilon to mu}
The next result, adapted from \cite{Hutter:04uaibook}, shows that if there is a good model
of the (unknown) environment in $C_D$, then Action-Conditional CTW will predict well.

\begin{thm}\label{conv to mu}
Let $\mu$ be the true environment, and $\Upsilon\equiv P^\epsilon_w$ the mixture environment model formed from (\ref{corr 3}).
The $\mu$-expected squared difference of $\mu$ and $\Upsilon$ is bounded as follows.
For all $n \in \mathbb{N}$, for all $a_{1:n}$,
\begin{gather}
\hspace{-2em} \sum_{k=1}^n \sum_{x_{<k}} \mu(x_{<k} \cdbar a_{<k}) \sum_{x_k} \biggl( \mu(x_k \cbar ax_{<k}a_k) - \Upsilon( x_k \cbar ax_{<k}a_k ) \biggr)^2 \notag \\
\hspace{0em} \leq \min_{M \in C_D} \,\biggl\{\, \Gamma_D(M) \ln 2 + \text{KL}( \mu( \cdot \cdbar a_{1:n}) \,\|\, \Pr( \cdot \cdbar M, a_{1:n})) \,\biggr\}, \label{ctw conv}
\end{gather}
where $\text{KL}(\cdot \,\|\, \cdot)$ is the KL divergence of two distributions.
\end{thm}

If the RHS of (\ref{ctw conv}) is finite over all $n$, then the sum on the LHS
can only be finite if $\Upsilon$ converges sufficiently fast to $\mu$.
If KL grows sublinear in $n$, then $\Upsilon$ still converges to $\mu$ (in a weaker Cesaro sense),
which is for instance the case for all $k$-order Markov and all stationary processes $\mu$.

\paradot{Overall Result}
Theorem~\ref{conv to mu} above in conjunction with \cite[Thm.5.36]{Hutter:04uaibook} imply $V_{\Upsilon}^m(h)$ converges to $V_\mu^m(h)$
as long as there exists a model in the model class that approximates the unknown environment $\mu$ well.
This, and the consistency (\ref{uctc}) of the \searchalg\ algorithm, imply that $\hat{V}_\Upsilon^m(h)$ converges to $V^m_{\mu}(h)$.
More detail can be found in \cite{Veness:arXiv0909.0801}.

\section{Experimental Results}

This section evaluates our approximate AIXI agent on a variety of test domains.
The Cheese Maze, 4x4 Grid and Extended Tiger domains are taken from the POMDP literature.
The TicTacToe domain comprises a repeated series of games against an opponent who moves randomly.
The Biased RockPaperScissor domain is described in \cite{farias07}, which involves the agent repeatedly playing RockPaperScissor against an exploitable opponent.
Two more challenging domains are included: Kuhn Poker \cite{hoehn05}, where the agent plays second against a Nash optimal player and a partially observable version of Pacman described in \cite{Veness:arXiv0909.0801}.
With the exception of Pacman, each domain has a known optimal solution.
Although our domains are modest, requiring the agent to learn the environment from scratch {\em significantly} increases the difficulty of each of these problems.

\begin{table}[hb!]
\begin{center}
\small
\begin{tabular}{|l|c|c|c|c|c|}
\hline
 Domain &  $|\cA|$ &  $|\cO|$ &  $\cA$ / $\cO$ / $\cR$ bits &  $D$ &  $m$ \\
\hline
 Cheese Maze &  4 &  16 &  2 / 4 / 5 &  96 &  8\\
 Tiger &  3 &  3 &  2 / 2 / 7 &  96 &  5\\
 4 $\times$ 4 Grid &  4 &  1 &  2 / 1 / 1 &  96 &  12\\
 TicTacToe &  9 &  19683 &  4 / 18 / 3 &  64 &  9\\
 Biased RPS &  3 &  3 &  2 / 2 / 2 &  32 &  4\\
 Kuhn Poker &  2 &  6 &  1 / 4 / 3 &  42 &  2\\
 Pacman &  4  &  65536 &  2 / 16 / 8 &  64 &  8\\
\hline
\end{tabular}
\vspace{-0.75em}
\caption{Parameter Configuration}
\label{table:domain_description}
\end{center}
\vspace{-1em}
\end{table}

Table \ref{table:domain_description} outlines the parameters used in each experiment.
The sizes of the action and observation spaces are given, along with the number of bits used to encode each space.
The context depth parameter $D$ specifies the maximal number of recent bits used by the Action-Conditional CTW prediction scheme.
The search horizon is given by the parameter $m$.
Larger $D$ and $m$ increase the capabilities of our agent, at the expense of linearly increasing computation time; our values represent an appropriate compromise between these two competing dimensions for each problem domain.

\begin{figure}[t!]
\begin{center}
\includegraphics[width=0.98\columnwidth]{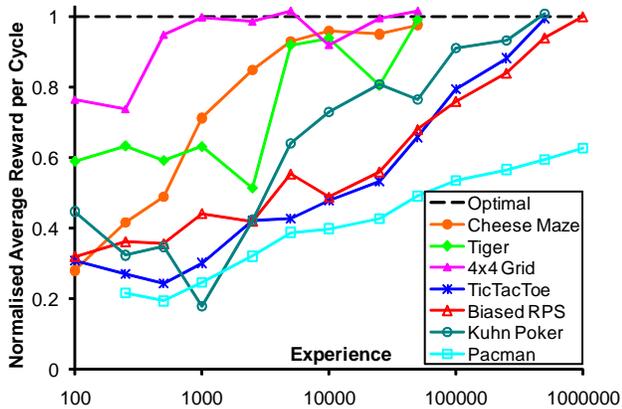}
\end{center}
\vspace{-2.35em}\caption{Learning scalability results}\label{fig:results}
\end{figure}

Figure \ref{fig:results} shows how the performance of the agent scales with experience, measured in terms of number of interaction cycles.
Experience was gathered by a decaying $\epsilon$-greedy policy, which chose randomly or used \searchalg.
The results are normalised with respect to the optimal average reward per time step, except in Pacman, where we normalised to an estimate.
Each data point was obtained by starting the agent with an amount of experience given by the $x$-axis and running it greedily for 2000 cycles.
The amount of search used for each problem domain, measured by the number of \searchalg\ simulations per cycle, is given in Table \ref{table:results}.
(The average search time per cycle is also given.)
The agent converges to optimality on all the test domains with known optimal values, and exhibits good scaling properties on our challenging Pacman variant.
Visual inspection\footnote{http://www.youtube.com/watch?v=RhQTWidQQ8U} of Pacman shows that the agent, whilst not playing perfectly, has already learnt a number of important concepts.

Table \ref{table:results} summarises the resources required for approximately optimal performance on our test domains.
Timing statistics were collected on an Intel dual 2.53Ghz Xeon.
Domains that included a planning component such as Tiger required more search.
Convergence was somewhat slower in TicTacToe; the main difficulty for the agent was learning not to lose the game immediately by playing an illegal move.
Most impressive was that the agent learnt to play an approximate best response strategy for Kuhn Poker, without knowing the rules of the game or the opponent's strategy.

\begin{table}
\begin{center}
\small
\begin{tabular}{|l|c|c|c|}
\hline
 Domain &  Experience &  Simulations &  Search Time\\
\hline
 Cheese Maze &  $5 \times 10^4$ &  500 &  0.9s\\
 Tiger &  $5 \times 10^4$ &  10000 &  10.8s\\
 4 $\times$ 4 Grid &  $2.5 \times 10^4$ &  1000 &  0.7s\\
 TicTacToe &  $5 \times 10^5$ &  5000 &  8.4s\\
 Biased RPS &  $1 \times 10^6$ &  10000 &  4.8s \\
 Kuhn Poker &  $5 \times 10^6$ &  3000 &  1.5s\\
\hline
\end{tabular}
\vspace{-0.5em}
\caption{Resources required for optimal performance}
\label{table:results}
\end{center}
\vspace{-1.9em}
\end{table}

\section{Related Work}

The BLHT algorithm \cite{suematsu99} is closely related to our work.
It uses symbol level PSTs for learning and an (unspecified) dynamic programming based algorithm for control.
BLHT uses the most probable model for prediction, whereas we use a mixture model, which admits a much stronger convergence result.
A further distinction is our usage of an Ockham prior instead of a uniform prior over PST models.

The Active-LZ \cite{farias07} algorithm combines a Lempel-Ziv based prediction scheme with dynamic programming for control to
produce an agent that is provably asymptotically optimal if the environment is $n$-Markov.
We implemented the Active-LZ test domain, Biased RPS, and compared against their published results.
Our agent was able to achieve optimal levels of performance within $10^6$ cycles; in contrast, Active-LZ was still suboptimal after $10^8$ cycles.

U-Tree \cite{mccallum96} is an online agent algorithm that attempts to discover a compact state representation from a raw stream of experience.
Each state is represented as the leaf of a suffix tree that maps history sequences to states.
As more experience is gathered, the state representation is refined according to a heuristic built around the Kolmogorov-Smirnov test.
This heuristic tries to limit the growth of the suffix tree to places that would allow for  better prediction of future reward.
Value Iteration is used at each time step to update the value function for the learned state representation, which is then used by the agent for action selection.

It is instructive to compare and contrast our AIXI approximation with the Active-LZ and U-Tree algorithms.
The small state space induced by U-Tree has the benefit of limiting the number of parameters that need to be estimated from data.
This has the potential to dramatically speed up the model-learning process.
In contrast, both Active-LZ and our approach require a number of parameters proportional to the number of distinct contexts.
This is one of the reasons why Active-LZ exhibits slow convergence in practice.
This problem is much less pronounced in our approach for two reasons.
First, the Ockham prior in CTW ensures that future predictions are dominated by PST structures that have seen enough data to be trustworthy.
Secondly, value function estimation is decoupled from the process of context estimation.
Thus it is reasonable to expect \searchalg\ to make good local decisions provided Action-Conditional CTW can predict well.
The downside however is that our approach requires search for action selection.
Although \searchalg\ is an anytime algorithm, in practice more computation is required per cycle compared to approaches like Active-LZ and U-Tree that act greedily with respect to an estimated global value function.

The U-Tree algorithm is well motivated, but unlike Active-LZ and our approach, it lacks theoretical performance guarantees.
It is possible for U-Tree to prematurely converge to a locally optimal state representation from which the heuristic splitting criterion can never recover.
Furthermore, the splitting heuristic contains a number of configuration options that can dramatically influence its performance \cite{mccallum96}.
This parameter sensitivity somewhat limits the algorithm's applicability to the general reinforcement learning problem.

Our work is also related to Bayesian Reinforcement Learning.
In model-based Bayesian RL \cite{pascal08,strens00}, a distribution over (PO)MDP parameters is maintained.
In contrast, we maintain an exact Bayesian mixture of PSTs.
The \searchalg\ algorithm shares similarities with Bayesian Sparse Sampling \cite{wang05}; the key differences are estimating the leaf node values with a rollout function and guiding the search with the UCB policy.

A more comprehensive discussion of related work can be found in \cite{Veness:arXiv0909.0801}.

\section{Limitations}

The main limitation of our current AIXI approximation is the restricted model class.
Our agent will perform poorly if the underlying environment cannot be predicted well by a PST of bounded depth.
Prohibitive amounts of experience will be required if a large PST model is needed for accurate prediction.
For example, it would be unrealistic to think that our current AIXI approximation could cope with real-world image or audio data.

The identification of \emph{efficient} and \emph{general} model classes that better approximate the AIXI ideal is an important area for future work.
Some preliminary ideas are explored in \cite{Veness:arXiv0909.0801}.

\section{Conclusion}

We have introduced the first computationally tractable approximation to the AIXI agent and shown that it provides a promising approach to the general reinforcement learning problem.
Investigating multi-alphabet CTW for prediction, parallelisation of \searchalg, further expansion of the model class (ideally, beyond variable-order Markov models) or more sophisticated rollout policies for \searchalg\ are exciting areas for future investigation.

\section{Acknowledgements}

This work received support from the Australian Research Council under grant DP0988049.
NICTA is funded by the Australian Government as represented by the Department of Broadband, Communications and the Digital Economy and the Australian Research Council through the ICT Centre of Excellence program.


\end{document}